\documentclass[letterpaper]{article} \usepackage[preprint]{aaai2027}  \usepackage[hyphens]{url}  \usepackage{graphicx}    \usepackage{natbib}  \usepackage{caption}   \usepackage{algorithm}
\usepackage{algorithmic}

\usepackage{newfloat}
\usepackage{listings}
\DeclareCaptionStyle{ruled}{labelfont=normalfont,labelsep=colon,strut=off} 
\floatstyle{ruled}
\newfloat{listing}{tb}{lst}{}
\floatname{listing}{Listing}

\usepackage{booktabs}

\title{Learning to Discretize: Diffusion-Based Adaptive Mesh with Spectral Guidance}
\author{
            Zixuan Shen\textsuperscript{\rm 1},
    Bingchuan Wang\textsuperscript{\rm 1},
    Zhi Wang\textsuperscript{\rm 2},
    Yong Wang\textsuperscript{\rm 1}
}
\affiliations{

    \textsuperscript{\rm 1}Central South University, Changsha, Hunan 410083, China\\
    \textsuperscript{\rm 2}Nanjing University, Nanjing, Jiangsu 210093, China\\

}

\begin{document}

\maketitle

\begin{abstract}
Most neural partial differential equation (PDE) surrogates learn how fields evolve after a grid has already been chosen. However, before any operator is applied, the grid has already determined how modeling capacity is allocated across space, resolution, and spectral bandwidth. We argue that this hidden design choice should itself be learnable, leading to a question different from standard operator learning: can a surrogate learn where resolution should exist before predicting field evolution? We formulate adaptive discretization as a physics-constrained conditional generation problem over valid mesh displacements. Similar to PDE field prediction, mesh generation requires producing structured outputs that satisfy physical and geometric constraints. The success of diffusion models in PDE field prediction suggests their potential for learning adaptive discretizations under similar structured constraints. This leads to a two-stage diffusion framework: Stage 1 learns an $r$-adaptive displacement mesh conditioned on the observed dynamics, while Stage 2 predicts the solution evolution from the mesh-informed representation. The mesh generator is regularized by physics-aware proxy channels, geometric validity constraints, and local spectral concentration so that adaptation remains physically interpretable and numerically legal. Across five PDE regimes, the results show that diffusion-based learned discretization is competitive with adaptive-mesh and reduced-order baselines, with particularly strong gains in regimes where fixed or handcrafted allocation is insufficient. The main conclusion is not that there exists a universal optimal mesh rule, but that \textbf{discretization should be learned in a regime-dependent manner}: different spatial and spectral structures favor different allocation behaviors. This reframes adaptive meshing for neural PDE solvers from a solver-specific heuristic into a generative representation-learning problem.

\end{abstract}
\section{Introduction}

\begin{figure}[!t]
\centering
\includegraphics[width=\columnwidth]{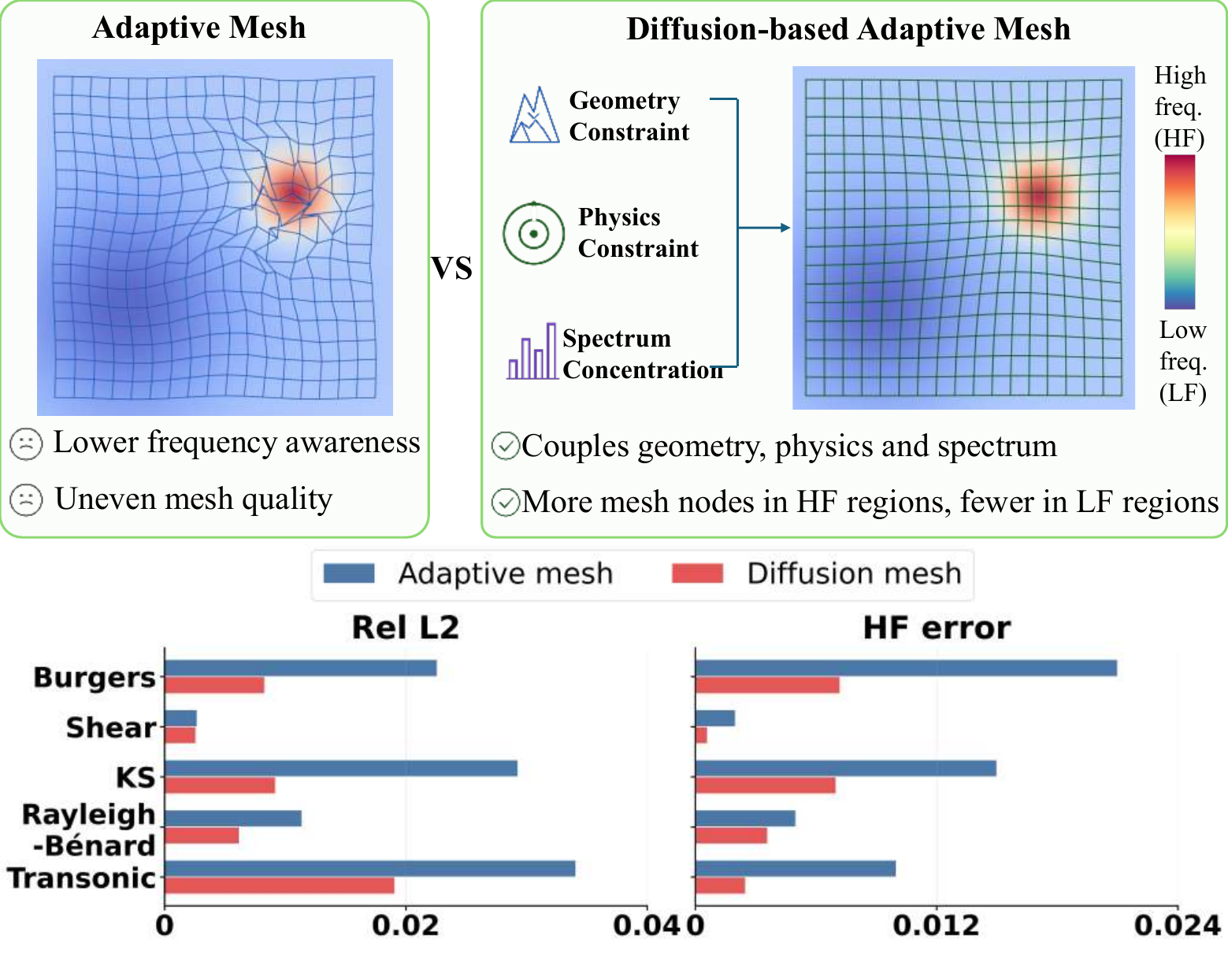}
\caption{Comparison between classical adaptive mesh and diffusion-based adaptive mesh (DAM). The proposed DAM formulates mesh generation as a conditional diffusion process that couples physics, geometry, and spectral information to generate frequency-aware displacement fields. This enables adaptive node allocation, concentrating resolution in high-frequency (HF) regions while preserving efficiency in low-frequency (LF) regions.}
\label{fig:Mesh Comparison}
\end{figure}

Partial differential equations (PDEs) are a central tool for modeling spatiotemporal phenomena, but modern neural surrogates still inherit one common assumption from classical pipelines: first choose the discretization, then learn the operator \cite{kovachki2023neural}. That assumption is convenient, yet it hides a representational bottleneck. \textbf{Discretization serves as the first representation layer, yet it is typically predefined and remains agnostic to the evolving solution state in most neural PDE solvers.} The grid determines where degrees of freedom are spent, which structures are resolved, and which frequencies survive before any learned operator is applied. This leads to a prior scientific question before operator design: can the representation itself be learned from the observed PDE state? We therefore view mesh displacement as a learnable representation policy: given the observed PDE history and a fixed node budget, the model should learn a valid coordinate deformation that reallocates degrees of freedom toward prediction-relevant structures. This policy is itself a structured prediction problem: like PDE field prediction, mesh generation must produce a coherent field under physical and geometric constraints. These similarities suggest that diffusion models, which have recently shown strong performance in PDE field prediction, may also be well-suited for mesh generation, where valid displacement fields are inferred from observed dynamics before solution prediction.

Recent years have brought notable progress in neural surrogates, including operator learning, spectral neural architectures, and adaptive discretization, but existing neural PDE frameworks still suffer from several important limitations. Operator methods such as DeepONet \cite{lu2021learning} provide resolution-agnostic mappings between function spaces; however, these frameworks often rely on a fixed discretization at training or inference time, which constrains their ability to concentrate resources. Spectral approaches like the Fourier Neural Operator \cite{li2020fourier} exploit global Fourier structure effectively, yet they typically assume regular lattices and may underperform when localized high-frequency structures dominate. Classical adaptive-meshing theory \cite{babuvvska1978error,berger1984adaptive,oden2001goal} supplies principled error estimators and moving-mesh techniques, but applying those tools in learned surrogates is nontrivial because estimator design and mesh updates are traditionally solver-specific.

These lines of work leave a gap than that accuracy alone suggests. Classical adaptivity often depends on solver-specific monitor functions or error estimators, while learned mesh baselines typically use graph message passing, sizing-field prediction, handcrafted refinement proxies, or single-pass coordinate regression. These mechanisms are useful, but they do not model mesh placement as a state-conditioned evolution process jointly coupled to the solution dynamics. What remains missing is a neural PDE formulation in which discretization is a first-class representation variable, optimized jointly with the operator. In that sense, operator learning has largely asked: \textbf{how should fields evolve on a grid?} We ask the prior question: \textbf{What grid should the model choose before predicting the solution, given physical and geometric constraints?}

\begin{figure*}[ht]
\centering
\includegraphics[width=0.94\textwidth]{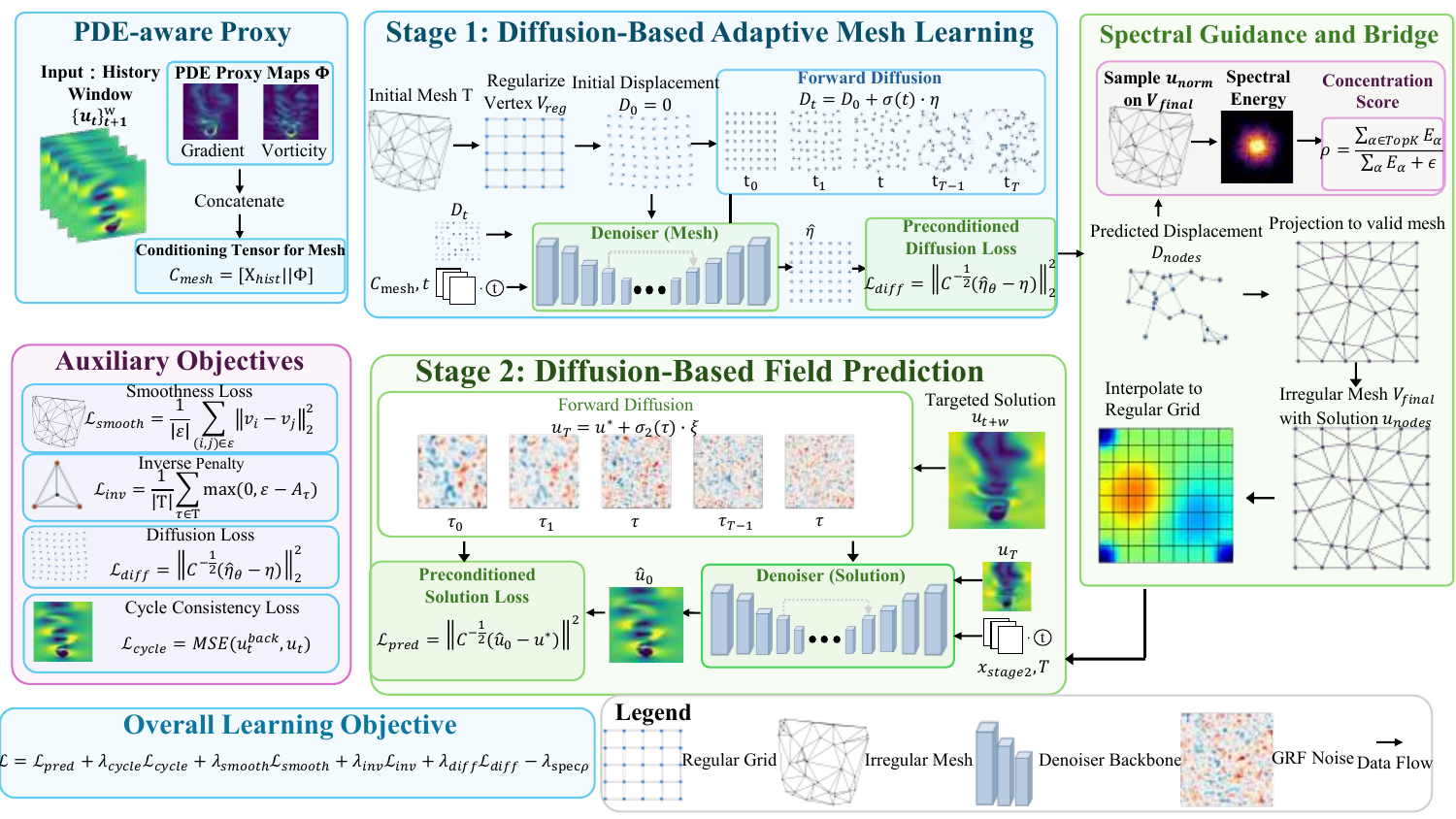}
\caption{Architecture overview of DAM, the proposed two-stage diffusion-based adaptive mesh framework for PDE field prediction. Stage 1 generates adaptive meshes through conditional diffusion in displacement space, conditioned on physical proxy maps and guided by local spectral concentration. The generated irregular mesh is bridged back to a tensor-compatible regular grid. Stage 2 performs solution-space diffusion for next-step PDE forecasting on the mesh-informed conditioning tensor, with CNN-based denoising used in both diffusion stages.}
\label{fig:architecture}
\end{figure*}

The effectiveness of the diffusion model for mesh generation is reflected in how it answers the preceding question: what grid should the model choose before evolution is predicted? In our formulation, this choice is not learned from geometry alone. The evolution process is coupled with geometric validity, physics-aware conditioning, spectral concentration, and downstream forecasting loss, so mesh placement is optimized as part of the predictive pipeline. Additionally, using diffusion models for both mesh generation and solution prediction further places the entire framework within a unified denoising paradigm, where both displacement and solution fields are modeled through iterative refinement under these constraints. The resulting mesh is therefore a structured displacement evolution from a reference lattice, not an unconstrained set of node perturbations. Figure 1 presents a comparison between adaptive discretization and DAM to illustrate the formulation: through this coupled objective, the model learns how resolution should move across spatial structures and frequency bands, improving multi-scale recovery by reallocating capacity between smooth low-frequency regions and localized high-frequency details before the solution model acts.

Motivated by this gap, the proposed DAM is an end-to-end two-stage framework that jointly learns adaptive discretization and PDE field prediction. In Stage 1, a conditional diffusion model acts as a mesh-adaptation operator, generating $r$-adaptive displacements from physics-aware context while proxy conditioning and geometric projection keep the mesh informative and valid. Learning in displacement space ties the generated mesh to a reference lattice, which makes identity initialization, bounded sampling, and validity projection natural. The mesh branch is trained through the forecasting objective that operates on the generated mesh, so a discretization is rewarded not merely for appearing adaptive, but for improving predictive performance. In Stage 2, the irregular mesh is bridged back to a regular tensor and coupled with a solution-evolution diffusion denoiser. A local spectral concentration objective links mesh allocation to the frequency structure of the observed field. 
To the best of our knowledge, this is the first attempt to formulate discretization as a conditional generative task. Our experiments demonstrate why this perspective matters: effective discretizations are strongly regime-dependent, varying with the spatial and spectral characteristics of the underlying PDE dynamics.

Our contributions are summarized as follows.
\begin{itemize}
\item \textbf{Coupled mesh adaptation and solution evolution:} We formulate PDE field prediction as a coupled learning problem in which a mesh-adaptation diffusion model first chooses where resolution should exist and a solution-evolution diffusion model then predicts the next field under that learned discretization.

\item \textbf{Diffusion-based $r$-adaptive discretization:} Instead of relying on graph-style message passing or a fixed sizing heuristic, we instantiate mesh adaptation as conditional denoising over displacement fields, with local spectral guidance and geometric projection keeping the generated meshes informative and legal.

\item \textbf{Physics-aware conditioning for mesh generation:} We compute causal proxy channels from the latest observed frame and concatenate them with the history tensor, so Stage 1 receives physically informed conditioning while the final discretization remains trained end-to-end through forecasting and spectral objectives.

\item \textbf{Evidence for a new design principle:} We empirically find that PDEs with different spatial and spectral characteristics favor different useful allocation patterns and mechanism choices, indicating that future neural PDE solvers should treat discretization as a regime-dependent learned representation rather than inherit one fixed rule.

\end{itemize}

\section{Related Work}
\textbf{Neural operators and frequency-aware learning.}
Neural operators form a practical route to resolution-agnostic surrogate models. DeepONet \cite{lu2021learning} and the broader neural operator program \cite{kovachki2023neural} show how mappings between function spaces can be learned. Fourier-based approaches such as the Fourier Neural Operator \cite{li2020fourier} and F-FNO \cite{lehmann20243d} place frequency structure at the core of the architecture, which improves recovery of oscillatory modes but often assumes a regular lattice. More recent variants broaden that design space in different ways: DRIFT-Net \cite{li2025drift} introduces spectral coupling, LoGlo-FNO \cite{kalimuthu2025loglo} combines local and global features, and Transolver++ \cite{luo2025transolver++} targets scalable PDE solving on large geometries. These designs advance accuracy; however, they rarely make mesh placement an explicit, learnable degree of freedom tied to the operator's loss. In short, operator advances give powerful inductive biases, yet they stop short of solving adaptive discretization jointly with prediction.

\textbf{Classical adaptive meshing and learning-based discretization.}
Adaptive meshing offers one of the most principled approaches to allocating resolution where it is needed most. Error estimators, moving-mesh methods, and goal-oriented adaptivity were established by works such as \cite{babuvvska1978error,berger1984adaptive,oden2001goal} and later practicalized in a posteriori techniques \cite{ainsworth1997posteriori,huang2010adaptive}. These methods deliver provable error control in many settings, but they typically require solver-specific estimators and handcrafted refinement rules. More recently, data-driven mesh strategies have reframed mesh design as a learning problem: neural \(r\)-adaptive finite element method (FEM) and refinement predictors \cite{aballay2025r,freymuth2025amber} learn parametric deformations, reinforcement-learning approaches \cite{thacher2025optimization} search for mesh policies, and learned mesh-movement techniques \cite{song2022m2n} adapt nodes based on observed dynamics. Work on graph- and point-based operators \cite{pfaff2020learning,qi2017pointnet,li2023fourier} shows that operator architectures can accommodate irregular domains, but integrating spectral objectives with mesh learning remains an open question. Thus, the mesh literature supplies strong tools, while learned methods offer flexibility, suggesting a natural opportunity to combine the strengths of both.

\textbf{Diffusion models for structured data generation.}
Diffusion and score-based models have matured into versatile generators for structured data. Foundational formulations \cite{song2019generative,ho2020denoising,song2020score} and developments in conditional guidance and latent diffusion \cite{ho2022classifier,hu2024anomalydiffusion,cai2026stabilizing} made controlled synthesis practical. Reduced-order PDE generators such as Text2PDE \cite{zhou2024text2pde}, Latent-FM \cite{li2025generative}, AROMA \cite{serrano2024aroma}, and LoLA \cite{rozet2026lost} show that generative modeling can be effective for compact PDE simulation. These methods mainly learn latent state evolution or reduced representations after discretization has been chosen. In contrast, our use of diffusion targets the discretization layer itself: the denoising process generates adaptive mesh displacements before the solution-evolution model predicts the future field.

\section{Method}

\subsection{Overview}
We study next-step PDE field prediction with learned discretization. Each sample provides a normalized history tensor and a target future state on the same physical domain. The proposed DAM couples two conditional diffusion processes. The first generates an $r$-adaptive mesh by denoising a displacement field, while a differentiable bridge maps the resulting irregular representation back to tensor form. The second predicts the next PDE state by denoising a solution field conditioned on this mesh-informed representation.

This structure separates two distinct generative questions that are often entangled in conventional predictors: where representational capacity should be allocated and what future state should be predicted under that representation. The former is addressed through mesh generation, while the latter is addressed through solution prediction. It is important because mesh ambiguity and solution uncertainty arise in different spaces and benefit from different inductive biases.

Thus, diffusion is used not only to generate solution fields, but also to generate the discretization used for prediction. Spectral guidance rewards informative resolution allocation, and geometric projection preserves mesh validity. Figure~2 summarizes the framework. This formulation keeps the problem close to standard forecasting while changing the objective being learned: before prediction begins, the model has already generated a resolution layout that determines which spatial and spectral information is represented. It also provides a natural mechanism for incorporating physical information, since physics-derived proxy channels can shape the generated mesh before solution prediction takes place.

\subsection{Diffusion-based Adaptive Mesh Learning}
We parameterize mesh generation in displacement space rather than absolute coordinates, so the diffusion model learns a deformation of a reference lattice. This keeps the sampling target bounded, preserves a natural identity initialization, and allows projection to enforce valid element orientation. Let $\mathbf{D}_0\in\mathrm{R}^{D\times H_m\times W_m}$ be the displacement field from a reference mesh, where $H_m$ and $W_m$ are mesh resolutions, and the two channels are horizontal/vertical displacement. With diffusion time $t\in[0,1]$, noise scale $\sigma(t)$, and Gaussian noise $\eta$, Stage 1 corrupts the displacement as:
\begin{equation}
\mathbf{D}_t=\mathbf{D}_0+\sigma(t)\eta,\qquad \eta\sim\mathcal{N}(\mathbf{0},\mathbf{I}),
\end{equation}
where $\mathbf{I}$ is the identity covariance. A denoiser $f_\theta$ predicts the injected noise from the noisy displacement, time, and mesh condition $\mathbf{C}_{\mathrm{mesh}}$:
\begin{equation}
\hat{\eta}_\theta=f_\theta(\mathbf{D}_t,t,\mathbf{C}_{\mathrm{mesh}}).
\end{equation}
The displacement prior is a Gaussian random field (GRF) with spectrum $\lambda(\mathbf{k})=(\alpha+\|\mathbf{k}\|^2)^{-s}$, where $\mathbf{k}$ denotes the frequency index, and $\alpha$ and $s$ control scale/smoothness. With $C^{-1/2}$ denoting the corresponding whitening operator, the diffusion loss is:
\begin{equation}
\mathcal{L}_{\mathrm{diff}}=\left\|C^{-1/2}(\hat{\eta}_\theta-\eta)\right\|_2^2.
\end{equation}
After denoising, the sampled node displacement $\mathbf{D}_{\mathrm{nodes}}$ is added to reference nodes $\mathbf{V}_{\mathrm{reg}}$ and projected by $\Pi$ onto the valid-mesh set:
\begin{equation}
\mathbf{V}=\Pi(\mathbf{V}_{\mathrm{reg}}+\mathbf{D}_{\mathrm{nodes}}),
\end{equation}
so learned deformation remains expressive while inversion is prevented in practice.

\subsection{Physics-Data Integrated Mesh Condition}
The mesh generator is conditioned on the reshaped history tensor $\mathbf{X}_{\mathrm{hist}}$ and a causal PDE-aware proxy $\Phi$ extracted only from the latest observed frame:
\begin{equation}
\mathbf{C}_{\mathrm{mesh}}=[\mathbf{X}_{\mathrm{hist}};\Phi],
\end{equation}
where $[\cdot;\cdot]$ denotes channel concatenation. Here $\Phi$ is a stack of PDE-aware proxy maps, such as gradient magnitude and vorticity magnitude. This causal conditioning tells Stage 1 where the current field is steep, rotating, compressive, or otherwise physically distinctive without exposing future targets.

As an example, for 2D fields with components $(u,v)$, the shared proxy channels physcially summarize local gradient magnitude and vorticity magnitude:
\begin{equation}
\phi_{\mathrm{grad}}=\sqrt{(\partial_xu)^2+(\partial_yu)^2+(\partial_xv)^2+(\partial_yv)^2+\varepsilon},
\end{equation}
\begin{equation}
\phi_{\mathrm{vort}}=|\partial_xv-\partial_yu|.
\end{equation}
Here $\phi_{\mathrm{grad}}$ is the gradient-magnitude channel, $\phi_{\mathrm{vort}}$ is the vorticity-magnitude channel, $\partial_x$ and $\partial_y$ are finite-difference derivatives, and $\varepsilon$ prevents numerical degeneracy. We do not require the same channel to dominate every dataset. The proxy stack instead gives Stage 1 a physically meaningful basis from which the diffusion sampler can learn dataset-specific corrections through the forecasting objective.

\subsection{Spectral Guidance and Bridge}
Purely spatial heuristics do not show whether capacity is allocated to operator-relevant frequencies. We compute a local spectral concentration score from the generated mesh and latest field. Let $E_a$ denote the local spectral energy at anchor $a$, and let $\mathrm{TopK}(E,k)$ be the set of the $k$ anchors with largest energy. We define:
\begin{equation}
\rho=\frac{\sum_{a\in\mathrm{TopK}(E,k)}E_a}{\sum_a E_a+\varepsilon},
\end{equation}
where $\varepsilon$ stabilizes the ratio. Maximizing $\rho$ rewards concentration on informative anchors without changing field scale, and the same module can append a local spectral map to Stage 2. The downstream stage does not operate directly on irregular nodes. The bridge reconstructs a regular tensor from the adaptive mesh and augments it with interpolation confidence, density, and local spectral cues, yielding the tensor $\mathbf{x}_{\mathrm{stage2}}$. The bridge is important because it lets the model exploit irregular sampling without abandoning tensorized prediction. It also carries discretization information beyond the reconstructed field, exposing density and local spectral cues to the downstream predictor.

\subsection{Diffusion-Based Field Prediction}
Given the tensor $\mathbf{x}_{\mathrm{stage2}}$, Stage 2 diffuses the normalized target solution $\mathbf{u}^*$ at diffusion time $\tau$ with noise scale $\sigma_2(\tau)$ and predicts the clean field:
\begin{equation}
\mathbf{u}_\tau=\mathbf{u}^*+\sigma_2(\tau)\xi,\qquad
\hat{\mathbf{u}}_0=g_\psi(\mathbf{u}_\tau,\tau,\mathbf{x}_{\mathrm{stage2}}).
\end{equation}
Here $\xi\sim\mathcal{N}(\mathbf{0},\mathbf{I})$, $g_\psi$ is the prediction denoiser, and $\hat{\mathbf{u}}_0$ is the predicted clean next field. Stage 2 therefore receives both a state estimate and compact cues describing how that estimate was produced. This is the point at which learned discretization becomes visible to the solution model; the mesh is not discarded after interpolation, but leaves density, confidence, and spectral traces in the conditioning tensor. For the PDE field prediction, supervision is imposed in the clean-solution space through a preconditioned metric:
\begin{equation}
\mathcal{L}_{\mathrm{pred}}=\left\|C_u^{-1/2}(\hat{\mathbf{u}}_0-\mathbf{u}^*)\right\|_2^2
\end{equation}
where $C_u^{-1/2}$ is the solution-space GRF preconditioner.

\subsection{Overall Learning Objective}

In summary, the overall learning objective is:
\begin{equation}
\begin{array}{rcl}
\mathcal{L} &=& \mathcal{L}_{\mathrm{pred}}+\lambda_{\mathrm{cycle}}\mathcal{L}_{\mathrm{cycle}}+\lambda_{\mathrm{smooth}}\mathcal{L}_{\mathrm{smooth}} \\
&& {}+\lambda_{\mathrm{inv}}\mathcal{L}_{\mathrm{inv}}+\lambda_{\mathrm{diff}}\mathcal{L}_{\mathrm{diff}}-\lambda_{\mathrm{spec}}\rho.
\end{array}
\end{equation}
Here $\mathcal{L}_{\mathrm{cycle}}$ measures the bridge reconstruction consistency, $\mathcal{L}_{\mathrm{smooth}}$ penalizes the neighboring-node displacement roughness, $\mathcal{L}_{\mathrm{inv}}$ discourages the signed-area collapse, and $\mathcal{L}_{\mathrm{diff}}$ is exactly the Stage 1 diffusion loss defined above. The auxiliary terms tie the learned discretization to geometric validity, mesh-diffusion guidance, and informative local spectra; exact formulations and implementation details are provided in the appendix. Notably, the objective is coupled: the mesh branch is optimized through the downstream forecasting task, so a discretization is rewarded only when it is both geometrically valid and beneficial for prediction.

Additionally, because Stage 1 predicts displacements rather than arbitrary coordinates, the reference mesh supplies stable topology, and projection acts on deformation magnitude rather than repairing connectivity after the fact. Smoothness suppresses oscillatory neighboring-node motion, while inversion loss discourages signed-area collapse during training. The diffusion model can therefore express nontrivial adaptive motion, but the learned mesh remains bridge-compatible.

\subsection{Training and Inference Path}
During training, the normalized history and active PDE-aware proxies form the Stage 1 condition. The mesh denoiser receives a noisy displacement field together with this context, while the clean displacement supervises the mesh diffusion branch. After projection, generated nodes pass through the bridge to form the Stage 2 conditioning tensor. Thus, the predictor sees mesh statistics induced by the current generator, not a frozen preprocessing artifact.

At inference time, the learned mesh is sampled before the next-step field is generated. The resulting control flow is intentionally causal:
\[
\mathbf{U}_{1:w}\rightarrow \mathbf{C}_{\mathrm{mesh}}\rightarrow \mathbf{V}\rightarrow \mathbf{x}_{\mathrm{stage2}}\rightarrow \hat{\mathbf{u}}_{t+w}.
\]
In this path, $\mathbf{U}_{1:w}$ is the observed history window, $\mathbf{C}_{\mathrm{mesh}}$ is the Stage 1 conditioning tensor, $\mathbf{V}$ is the generated adaptive mesh, $\mathbf{x}_{\mathrm{stage2}}$ is the bridged regular-grid tensor, and $\hat{\mathbf{u}}_{t+w}$ is the next-state prediction. The arrow sequence also specifies the causal dependency at test time: the mesh is sampled before the solution denoiser runs, and the denoiser only receives the bridged representation induced by that sampled mesh. Only the observed history is used to decide the discretization. This design prevents the adaptive mesh from becoming an implicit oracle. Proxy features are computed solely from observed states, and teacher displacements are used only as training supervision rather than inference-time inputs. Consequently, any performance gain must arise from learning a useful discretization policy from available dynamics, rather than from access to future residuals, gradients, or target information.

\section{Experiments}

\subsection{Experimental Setup}

We evaluate on PDE benchmarks that exhibit distinct spatial and spectral structures, including transport-dominated dynamics, shear-driven instabilities, chaotic evolution, buoyancy-driven convection, and compressible transonic flows. These benchmarks are useful not only because they test prediction accuracy, but because they let us ask whether the same discretization behavior is suitable across distinct regimes. Table~\ref{tab:results} first compares adaptive-discretization baselines adapted to our forecasting protocol, including ATS~\cite{fayyaz2022ats}, AMR-MGNN~\cite{perera2024multiscale}, AMBER~\cite{freymuth2025amber}, and wLMR~\cite{ozgenxian2020wavelet}, because they directly test alternative ways of allocating resolution before prediction. We also include Text2PDE~\cite{zhou2024text2pde} as a reduced-order generative baseline: although it is not an adaptive-mesh method, it is relevant because both approaches reduce the effective complexity faced by generative PDE field prediction by changing the representation, with Text2PDE using latent compression and DAM using mesh-based allocation before diffusion-based solution prediction. All methods use the same input history and prediction horizon. In all mesh experiments, the displacement field uses two coordinate channels; equivalently, $D=2$ in $\mathbf{D}_0\in\mathrm{R}^{D\times H_m\times W_m}$. The denoising network in both stages of DAM is implemented as a convolutional neural network (CNN). We report field and spectral errors in Table~\ref{tab:results}, and use the remaining analyses to isolate what is learned by the discretization module rather than treating raw prediction score as the sole target.

\subsection{Predictive Validity of Learned Discretization}

\begin{table*}[ht]
\centering\small
\setlength{\tabcolsep}{1mm}
\renewcommand{\arraystretch}{1.12}
\begin{tabular}{@{}l l c c c c c c c c c c@{}}
\hline
Dataset & Metric & Reduced & \multicolumn{4}{c}{Adaptive discretization} & \multicolumn{5}{c}{DAM-family} \\
\cmidrule(lr){4-7}\cmidrule(lr){8-12}
 &  & Text2PDE & ATS & AMR-MGNN & AMBER & wLMR & 
 \begin{tabular}{@{}c@{}}Diff-\\only\end{tabular} &
 \begin{tabular}{@{}c@{}}Classic+\\Diff\end{tabular} &
 \begin{tabular}{@{}c@{}}Diff+\\CNN\end{tabular} &
 \begin{tabular}{@{}c@{}}Diff+\\FNO\end{tabular} & DAM \\
\hline
Burgers & Rel L2 & 0.0564 & 0.0088 & 0.0150 & 0.0371 & 0.0294 & 0.2032 & 0.0156 & 0.0132 & $\mathbf{0.0075}$ & $0.0083\pm0.0041$ \\
 & RMSE & 0.0163 & 0.0027 & 0.0046 & 0.0108 & 0.0087 & 0.0612 & 0.0045 & 0.0039 & $\mathbf{0.0022}$ & $0.0024\pm0.0012$ \\
 & Spec L2 & 0.0465 & 0.0136 & 0.0164 & 0.0245 & 0.0305 & 0.0338 & 0.0192 & 0.0069 & $\mathbf{0.0027}$ & $0.0072\pm0.0056$ \\
\hline
Shear & Rel L2 & 0.0801 & 0.0032 & 0.0047 & $\mathbf{0.0013}$ & 0.0017 & 0.0553 & 0.0099 & 0.0024 & 0.0030 & $0.0026\pm0.0005$ \\
 & RMSE & 0.0400 & 0.0016 & 0.0024 & $\mathbf{0.0007}$ & 0.0008 & 0.0277 & 0.0050 & 0.0012 & 0.0015 & $0.0013\pm0.0002$ \\
 & Spec L2 & 0.0286 & 0.0033 & 0.0034 & 0.0009 & 0.0010 & 0.0332 & 0.0138 & $\mathbf{0.0002}$ & 0.0052 & $0.0006\pm0.0003$ \\
\hline
KS & Rel L2 & 0.0553 & 0.0418 & 0.0295 & 0.0260 & 0.0200 & 0.1322 & 0.0198 & 0.0126 & 0.0229 & $\mathbf{0.0092\pm0.0009}$ \\
 & RMSE & 0.0181 & 0.0095 & 0.0066 & 0.0056 & 0.0045 & 0.0300 & 0.0040 & 0.0028 & 0.0047 & $\mathbf{0.0020\pm0.0002}$ \\
 & Spec L2 & 0.0718 & 0.0119 & 0.0228 & 0.0135 & 0.0123 & 0.0413 & 0.0283 & $\mathbf{0.0039}$ & 0.0099 & $0.0070\pm0.0033$ \\
\hline
Rayleigh & Rel L2 & 0.0831 & 0.0112 & 0.0154 & 0.0072 & 0.0117 & 0.3780 & 0.0380 & 0.0065 & $\mathbf{0.0060}$ & $0.0062\pm0.0011$ \\
 -B\'{e}nard& RMSE & 0.0109 & 0.0015 & 0.0021 & 0.0009 & 0.0016 & 0.0620 & 0.0052 & 0.0009 & $\mathbf{0.0008}$ & $\mathbf{0.0008\pm0.0001}$ \\
 & Spec L2 & 0.0207 & 0.0062 & 0.0072 & $\mathbf{0.0024}$ & 0.0040 & 0.4617 & 0.0199 & 0.0060 & 0.0025 & $0.0036\pm0.0013$ \\
\hline
Transonic & Rel L2 & 0.1164 & 0.0252 & 0.0254 & 0.0480 & 0.0376 & 0.1910 & 0.0202 & 0.0205 & 0.0259 & $\mathbf{0.0191\pm0.0009}$ \\
 & RMSE & 0.0602 & 0.0132 & 0.0132 & 0.0253 & 0.0197 & 0.0996 & 0.0105 & 0.0107 & 0.0135 & $\mathbf{0.0099\pm0.0005}$ \\
 & Spec L2 & 0.0257 & 0.0062 & 0.0147 & 0.0110 & 0.0081 & 0.0047 & 0.0023 & 0.0047 & $\mathbf{0.0020}$ & $0.0025\pm0.0019$ \\
\hline
\end{tabular}
\caption{Main prediction performance (lower is better), ordered roughly from easier to harder regimes. Text2PDE is listed as a reduced-order generative baseline; ATS, AMR-MGNN, AMBER, and wLMR are adaptive discretization methods. The DAM-family variants isolate discretization and predictor choices: Diff-only removes mesh conditioning, Classic+Diff uses a classical adaptive mesh with the same CNN-based prediction denoiser, Diff+CNN uses the DAM mesh with a pure CNN predictor, and Diff+FNO uses the DAM mesh with a pure FNO predictor. Here Diff denotes the CNN-based prediction denoiser. DAM reports the two-stage diffusion model with CNN denoisers in both mesh and prediction stage.}
\label{tab:results}
\end{table*}

Our first question is a fundamental one: can discretization become learnable without making the surrogate unusable as a predictor? Table~\ref{tab:results} answers this question from two perspectives. Against external adaptive-discretization methods, DAM achieves the lowest field errors on Kuramoto-Sivashinsky (KS) and Transonic equations, remains competitive on Burgers and Rayleigh-B\'{e}nard, and is most clearly challenged by Periodic shear, where the field is smooth and nearly separable enough for AMBER and wLMR to exploit simple sizing rules. This is consistent with our claim: learned discretization is not a universal replacement for every grid heuristic, but a regime-dependent representation policy whose value appears when allocation cannot be captured by a single handcrafted rule.

The DAM-family variants then separate mesh quality from predictor capacity. DAM improves over Classic+Diff on KS, Rayleigh-B\'{e}nard, and Transonic, indicating that the learned displacement policy contributes beyond a handcrafted adaptive rule. Diff+CNN and Diff+FNO remain competitive in several regimes, showing that the learned mesh is not tied to one particular prediction backend, but the full two-stage diffusion path is usually stronger. Diff-only performs poorly in harder regimes, ruling out the interpretation that the gains come only from the prediction denoiser. These comparisons establish the empirical role of learned discretization; the following analyses ask what spatial and spectral allocation the mesh learns.

\subsection{Spectral and Regime Analysis}
Table~\ref{tab:results} quantifies the predictive gains of learned adaptation, while Figure~\ref{fig:prediction_heatmap} in the Appendix illustrates how those gains manifest in the predicted fields across different regimes. The visual comparison shows that the benefit is not simply a uniform increase in spatial resolution: in smooth or low-dimensional systems the learned mesh mostly preserves the dominant structure, while in convection- and shock-sensitive regimes it must allocate capacity to localized patterns. Text2PDE captures coarse field layout but tends to smooth localized structures; AMR-style baselines often recover the global field but differ in local sharpness; our diffusion mesh is the most useful when structure must be preserved without relying on a single hand-designed refinement indicator. This motivates the targeted spectral and mesh-quality audits below rather than a claim of one universal refinement rule.\vspace{-0.05cm}

\begin{figure}[!t]
\centering
\includegraphics[width=\columnwidth]{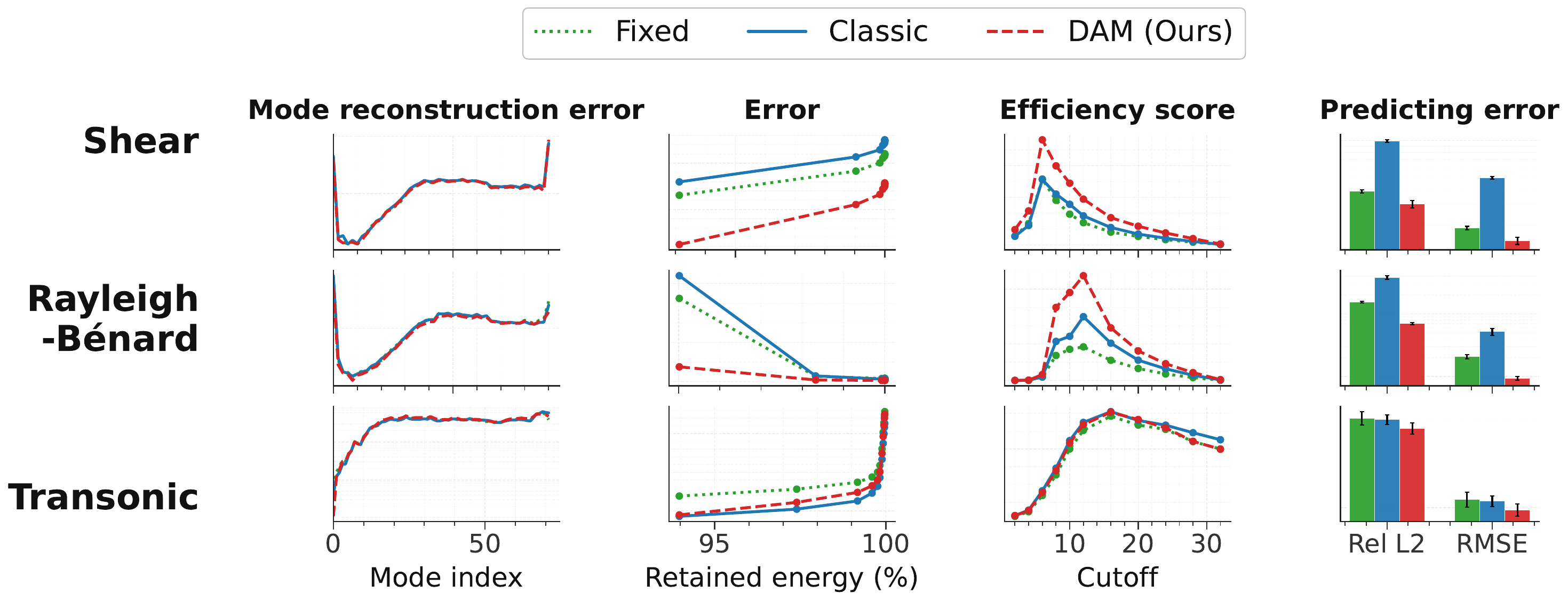}
\caption{Spectral representation under fixed-grid, classical adaptive, and DAM (Ours) discretization: Fourier-mode error, retained-energy error, efficiency score, and predicting error at resolution of $N=64$.}
\label{fig:spectral_concentration}
\end{figure}

\begin{figure*}[!t]
\centering
\includegraphics[width=0.98\textwidth]{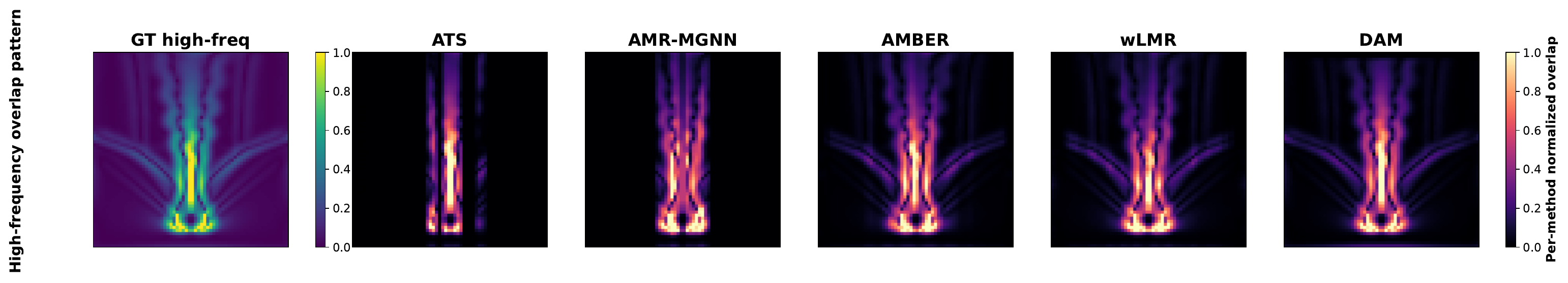}
\caption{Transonic spectral allocation. DAM covers the wake and high-frequency structures more comprehensively than ATS and AMR-MGNN, while allocating density in a more targeted manner than AMBER and wLMR.}
\label{fig:transonic_density_overlap}
\end{figure*}

\subsection{Ablation on Discretization Learning Mechanisms}
Because different components affect different stages of the pipeline, we use targeted diagnostics rather than a single aggregate ablation. We focus on the 2D benchmarks, where spatial and spectral allocation patterns are clearer than the simpler 1D cases, providing a cleaner test of mesh diffusion, spectral guidance, and proxy conditioning.

\subsubsection{Effect of Diffusion-Based Mesh Learning}

We first compare fixed, classical adaptive, and learned discretization to test whether diffusion-based mesh generation changes the representation rather than merely adding model capacity. Figure~\ref{fig:spectral_concentration} summarizes the results on the three 2D regimes. In Periodic Shear, all methods achieve similar spectral reconstructions, but DAM provides a better efficiency-accuracy trade-off and lower forecasting error. In Rayleigh-B\'{e}nard, DAM consistently reduces retained-energy, reconstruction, and forecasting errors, indicating a more informative allocation of resolution. In Transonic flow, the spectral metrics are closer across methods, yet DAM still achieves the lowest prediction error, suggesting that the learned mesh better preserves localized flow structures. These results indicate that the mesh diffusion branch does more than refine the grid: by generating state-dependent displacement fields, it changes how spatial and spectral information is allocated. Although initialized from a classical adaptive teacher during training, the learned policy is further shaped by forecasting and spectral objectives, leading to discretizations that are tailored to the downstream task.

\begin{figure}[!t]
\centering
\includegraphics[width=\columnwidth]{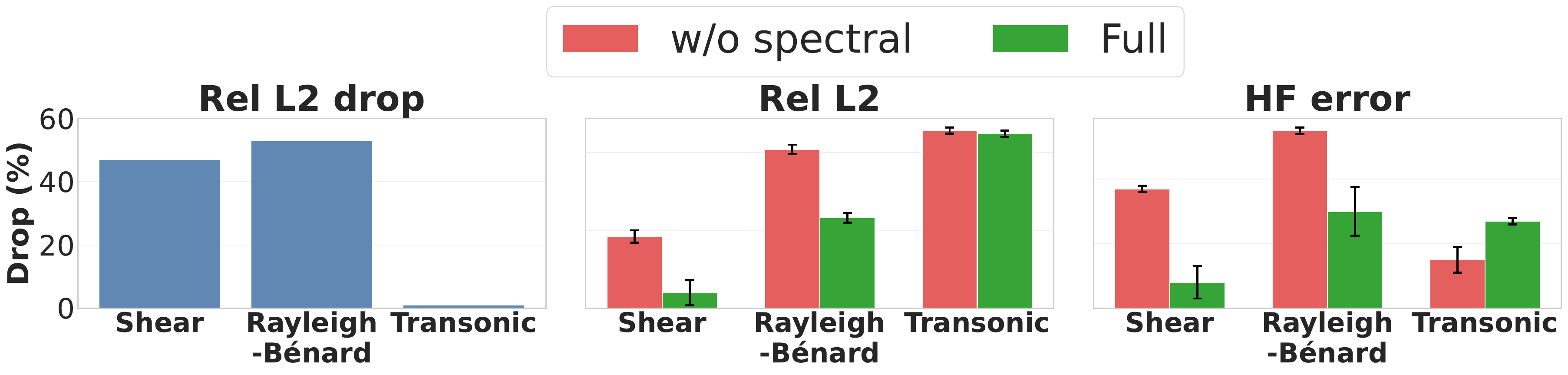}
\caption{Spectral-concentration ablation on 2D regimes. Positive drop indicates Rel L2 improvement from enabling $\lambda_{\mathrm{spec}}$.}
\label{fig:spectral_ablation}
\end{figure}

\subsubsection{Effect of Spectral Concentration Objective}
Building on the previous spectral analysis, we use Transonic to inspect where sampling density is placed because its low- and high-frequency structures are visually separable.

Figure~\ref{fig:transonic_density_overlap} shows that ATS and AMR-MGNN concentrate sampling narrowly around a few vertical bands, while DAM covers the body wake and surrounding high-frequency structures more completely. Compared with AMBER and wLMR, DAM achieves similar spatial coverage while allocating resolution in a more targeted manner around high-frequency regions, which is consistent with the lower Transonic error in Table~\ref{tab:results}. This supports the need for spectral guidance: the learned mesh is not only denser, but better aligned with the frequency structures that matter for prediction.

Figure~\ref{fig:spectral_ablation} compares the full model with an otherwise identical run using $\lambda_{\mathrm{spec}}=0$. The positive drop reports the Rel L2 improvement from enabling spectral concentration. The objective substantially lowers high-frequency error on Periodic shear and Rayleigh-B\'{e}nard, while Transonic shows a smaller but prediction-relevant trade-off. Overall, the spectral term changes what the mesh is rewarded to preserve, supporting the interpretation that DAM learns a frequency-aware allocation policy rather than only imitating a spatial adaptive prior.

\begin{figure}[!t]
\centering
\includegraphics[width=\columnwidth]{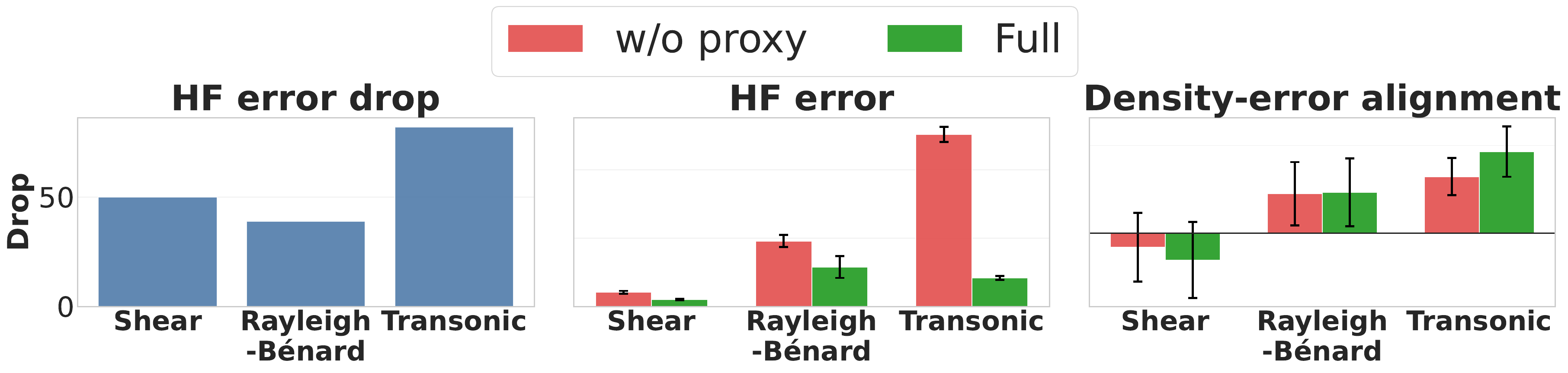}
\caption{PDE-aware proxy ablation on 2D regimes: high-frequency (HF) error drop, high-frequency error comparison, and density-error hotspot alignment.}
\label{fig:proxy_ablation}
\end{figure}

\subsubsection{Effect of PDE-aware Proxy Conditioning}
PDE-aware proxy conditioning is used only as a Stage 1 mesh input. It is computed from the latest observed frame and therefore cannot leak target information into the mesh policy. The ablation asks whether explicit physical cues change the learned allocation beyond what prediction loss and geometric regularization already provide. Figure~\ref{fig:proxy_ablation} shows that enabling the proxy reduces the high-frequency error diagnostic across the three 2D regimes. The alignment panel is interpreted more carefully: the effect is strongest when localized structure is meaningful, especially in Transonic flow, while smoother globally organized regimes show weaker correlation shifts. The proxy is therefore not presented as a universal raw-error trick; its role is to provide causal and physically interpretable conditioning for regime-dependent learned discretization.

\section{Conclusion}

We introduced DAM, a two-stage diffusion framework for jointly learning adaptive discretization and PDE field prediction. Instead of using a prescribed mesh, DAM denoises a displacement field to generate a state-dependent mesh, then predicts on the learned discretization. Across five PDE benchmarks, DAM improves prediction accuracy while maintaining geometrically valid meshes. The learned discretization also generalizes across nearby resolutions and benefits from geometry-, physics-, and frequency-aware conditioning, showing that it captures structures beyond handcrafted refinement heuristics. More broadly, these results suggest that adaptive discretization should be treated as a learnable representation, not fixed preprocessing, offering a new path toward joint mesh generation and neural PDE learning.

\bibliography{aaai2027}

\onecolumn
\appendix
\section{Appendix}
\subsection{Notation}
\begin{table}[ht]
\centering\renewcommand{\arraystretch}{1.08}
\begin{tabular}{p{0.24\textwidth} p{0.68\textwidth}}
\hline
Symbol & Meaning \\
\hline
$H,W$ & Spatial resolution of the regular PDE grid \\
$H_m,W_m$ & Resolution of the reference mesh used in Stage 1 \\
$w$ & Length of the observed history window \\
$\mathbf{u}_t=(u,v)$ & Two-channel PDE state at time $t$ \\
$\mathbf{u}^*$ & Normalized next-step target \\
$\mathbf{U}_{1:w}^{\mathrm{norm}}$ & Normalized history sequence used as model input \\
$\mathbf{X}_{\mathrm{hist}}$ & History sequence reshaped into $2w$ conditioning channels \\
$\Phi$ & PDE-aware proxy tensor built from the latest observed frame only \\
$\phi_{\mathrm{grad}},\phi_{\mathrm{vort}},\phi_{\mathrm{lap}}$ & Shared proxy channels for gradient magnitude, vorticity magnitude, and Laplacian magnitude \\
$\mathbf{C}_{\mathrm{mesh}}$ & Stage 1 conditioning tensor $[\mathbf{X}_{\mathrm{hist}};\Phi]$ \\
$\mathbf{D}_0,\mathbf{D}_t$ & Clean and noisy mesh-displacement fields in Stage 1 diffusion \\
$\eta,\sigma(t)$ & Stage 1 Gaussian noise and noise scale at diffusion time $t$ \\
$\mathbf{V}_{\mathrm{reg}},\mathbf{V}$ & Reference mesh nodes and projected adaptive mesh nodes \\
$\Pi$ & Geometry projection operator enforcing valid mesh deformation \\
$\mathcal{A},a$ & Local spectral anchor set and an individual anchor \\
$\omega_{n,a}$ & Soft assignment weight from mesh node $n$ to anchor $a$ \\
$c_{a,m},E_a$ & Local Fourier coefficient and anchor energy for spectral mode $m$ \\
$k,M$ & Number of selected top-energy anchors and number of retained local Fourier modes \\
$\rho$ & Local spectral concentration score \\
$\mathbf{x}_{\mathrm{stage2}}$ & Bridged regular-grid conditioning tensor for Stage 2 \\
$\mathbf{u}_\tau,\xi,\sigma_2(\tau)$ & Noisy Stage 2 solution, Gaussian noise, and Stage 2 noise scale at diffusion time $\tau$ \\
$C^{-1/2},C_u^{-1/2}$ & GRF-derived frequency preconditioners for mesh and prediction diffusion losses \\
$\lambda(\mathbf{k}),\alpha,s$ & GRF spectral density and its scale/smoothness hyperparameters \\
$\mathcal{E},\mathcal{T}$ & Mesh edge set and triangle set \\
$A_\tau$ & Signed area of triangle $\tau$ \\
$\lambda_{\mathrm{cycle}},\lambda_{\mathrm{smooth}},\lambda_{\mathrm{inv}}$ & Scalar weights of the cycle, smoothness, and inversion terms \\
$\lambda_{\mathrm{diff}},\lambda_{\mathrm{spec}}$ & Scalar weights of the mesh diffusion and spectral-concentration terms \\
$\mathcal{L}_{\mathrm{pred}},\mathcal{L}_{\mathrm{cycle}},\mathcal{L}_{\mathrm{smooth}}$ & Prediction, reconstruction-cycle, and mesh-smoothness losses \\
$\mathcal{L}_{\mathrm{inv}},\mathcal{L}_{\mathrm{diff}}$ & Inversion and mesh diffusion losses \\
$\mathrm{max\_disp\_ratio}$ & Maximum Displacement Ratio \\
\hline
\end{tabular}
\caption{Notation used throughout the paper.}
\label{tab:notation}
\end{table}

\subsection{Theoretical Notes}
\subsubsection{Geometry preservation.}
The learned mesh is parameterized as a displacement of a valid reference grid rather than as unconstrained coordinates. Before use, the raw displacement is clipped by $\mathrm{max\_disp\_ratio}$ and passed through a bisection-based projection that repeatedly rescales the deformation until every triangle area remains above a positive threshold. Because the reference triangulation is valid and the projection accepts only deformations with positive signed area, the returned mesh is inversion-free by construction up to numerical tolerance. The empirical minimum-Jacobian and inversion-rate results in the paper verify that the implementation follows this guarantee in practice.

\subsubsection{Why GRF preconditioning is appropriate.}
Let a displacement field follow a GRF prior with spectral density $\lambda(\mathbf{k})=(\alpha+\|\mathbf{k}\|^2)^{-s}$. Whitening the denoising residual by $C^{-1/2}$ yields the norm:
\begin{equation}
\|C^{-1/2}e\|_2^2=\sum_{\mathbf{k}}(\alpha+\|\mathbf{k}\|^2)^s |\hat e(\mathbf{k})|^2,
\end{equation}
which penalizes high-frequency deformation errors more strongly than low-frequency drift. This matches the modeling role of a mesh displacement: large-scale motion can be useful, whereas spurious high-frequency oscillation rapidly damages element quality.

\subsection{Detailed Method and Implementation Notes}\label{app:method_details}
\subsubsection{PDE-aware proxy construction.}
The implementation builds proxy channels from the latest normalized frame only, avoiding future leakage. Periodic finite differences provide
\begin{equation}
\phi_{\mathrm{grad}}=\sqrt{(\partial_xu)^2+(\partial_yu)^2+(\partial_xv)^2+(\partial_yv)^2+\varepsilon},
\end{equation}
\begin{equation}
\phi_{\mathrm{vort}}=|\partial_xv-\partial_yu|,
\end{equation}
and, when enabled, $\phi_{\mathrm{lap}}=|\Delta u|+|\Delta v|$. The current five-dataset configuration uses the same causal proxy principle across datasets: proxy channels are computed only from the observed history, then concatenated into the Stage~1 mesh-conditioning tensor. In the main configuration, $\phi_{\mathrm{lap}}$ is disabled, so $\Phi$ remains a compact stack of local derivative-based cues rather than a target-dependent signal.

\subsubsection{Local spectral concentration.}
Given anchor locations $\mathcal{A}$, node-to-anchor weights are defined by a softmax over squared Euclidean distances,
\begin{equation}
\omega_{n,a}=\mathrm{softmax}_a\!\left(-\frac{\|\mathbf{v}_n-\mathbf{a}\|_2^2}{\sigma^2}\right).
\end{equation}
The implementation projects the latest field onto a compact Fourier basis, forms anchor energies $E_a=M^{-1}\sum_m |c_{a,m}|^2$, and computes
\begin{equation}
\rho=\frac{\sum_{a\in\mathrm{TopK}(E,k)}E_a}{\sum_aE_a+\varepsilon}.
\end{equation}
The ratio is bounded in $[0,1]$. Maximizing $\rho$ therefore cannot increase total spectral energy arbitrarily; it only changes how energy is concentrated across anchors. The objective encourages the mesh to allocate resolution toward the most informative local spectral regions while remaining agnostic to the absolute field scale.

\subsubsection{Bridge and geometry projection.}
The bridge constructs a regularized grid tensor from mesh nodes through differentiable interpolation and concatenates confidence, density, and optional spectral channels. For stability, sampled displacement fields are first clamped by $\mathrm{max\_disp\_ratio}$ relative to the reference spacing and then passed through the projection operator $\Pi$, which repeatedly rescales the deformation until all signed triangle areas are positive. This mirrors the implementation used before every learned-mesh forward pass.

\subsubsection{Exact auxiliary losses.}
For mesh nodes $\mathbf{v}_i$ and graph edges $\mathcal{E}$,
\begin{equation}
\mathcal{L}_{\mathrm{smooth}}=\frac{1}{|\mathcal{E}|}\sum_{(i,j)\in\mathcal{E}}\|\mathbf{v}_i-\mathbf{v}_j\|_2^2,
\end{equation}
while for signed triangle areas $A_\tau$,
\begin{equation}
\mathcal{L}_{\mathrm{inv}}
=
\frac{1}{|\mathcal{T}|}
\sum_{\tau \in \mathcal{T}}
\max\left(0,\epsilon - A_{\tau}\right),
\qquad
\epsilon = 10^{-10}.
\end{equation}
The implementation also uses the cycle term between bridged reconstruction and latest context frame plus a mesh diffusion term for Stage 1 displacement supervision. The objective written in the main text therefore matches the code path used for the reported learned-mesh runs.

\begin{table*}[htbp]
\centering\small\renewcommand{\arraystretch}{1.12}
\begin{tabular}{l c c c c c p{0.29\textwidth}}
\hline
Dataset & Epochs & Steps/epoch & Batch & Train mesh & Test mesh & Key weights \\
\hline
Burgers & 200 & 100 & 8 & 64 & 64 & $\lambda_{\rm spec}=0.05$, $\lambda_{\rm cycle}=0.20$, max\_disp \_ratio $=0.25$ \\
Shear & 300 & 100 & 8 & 64 & 64 & $\lambda_{\rm spec}=0.05$, $\lambda_{\rm cycle}=0.1$ \\
KS & 200 & 100 & 8 & 64 & 64 & $\lambda_{\rm spec}=0.05$, $\lambda_{\rm cycle}=0.1$ \\
Rayleigh-B\'{e}nard & 300 & 100 & 8 & 64 & 64 & $\lambda_{\rm spec}=0.05$, $\lambda_{\rm cycle}=0.1$ \\
Transonic & 300 & 100 & 8 & 64 & 64 & $\lambda_{\rm spec}=0.05$, $\lambda_{\rm cycle}=0.1$ \\
\hline
\end{tabular}
\caption{Dataset-wise training parameters for DAM (Ours). Main Table~\ref{tab:results} uses seeds $\{1,2,42\}$. Shared parameters are history window $w=8$, horizon $=1$, learning rates $2\times10^{-4}$ for both mesh and solution branches, batch size 8, and $\lambda_{\mathrm{inv}}=1.0$. Unless otherwise noted, $\lambda_{\mathrm{smooth}}=10^{-4}$, mesh diffusion loss weight $=0.2$, and $\mathrm{max\_disp\_ratio}=0.45$; Burgers uses the optimized setting $\lambda_{\mathrm{smooth}}=2\times10^{-4}$ and mesh diffusion loss weight $=0.15$.}
\label{tab:training_details}
\end{table*}

\subsection{Training Details}

The mesh branch and downstream predictor are trained jointly, so the reported results reflect end-to-end optimization rather than a separately tuned mesh generator. The complete dataset-specific training configurations, including optimization hyperparameters and loss weights are summarized in Table 3.

\subsection{Hardware and Software Environment}
All experiments reported in this paper were executed on a single NVIDIA GeForce RTX 4070 Laptop GPU with 8\,GB memory (8188\,MiB reported by the driver; driver version 581.83). The active training environment used PyTorch 2.5.1 with CUDA 12.1. All runtime numbers in the paper were measured on this hardware rather than copied from vendor specifications. This disclosure is included to make the efficiency results interpretable, especially because diffusion-based mesh sampling adds non-negligible latency relative to a fixed-grid path.

\subsection{Runtime Cost of Learned Discretization}
Table~\ref{tab:runtime_cost} reports the measured total inference time of DAM at resolution of $N=64$. We report the total solution-generation path rather than separating solver and mesh-generation entries in this table; the mesh-generation component is analyzed separately in the Pareto figure below.

\begin{table}[htbp]
\centering\renewcommand{\arraystretch}{1.12}
\begin{tabular}{l c}
\hline
Dataset & DAM total time ms \\
\hline
Burgers & 315.1 \\
Shear & 145.6 \\
KS & 126.1 \\
Rayleigh-B\'{e}nard & 231.2 \\
Transonic & 790.1 \\
\hline
\end{tabular}
\caption{Total inference time of DAM (Ours) at resolution of $N=64$ in milliseconds per sample. The reported time corresponds to the full solution-generation path used during evaluation.}
\label{tab:runtime_cost}
\end{table}

DAM currently generates mesh displacements through an iterative denoising process, so the mesh-generation module itself leaves room for acceleration. To test whether this part of the framework can be replaced by a faster generative mechanism, we evaluate a Burgers case in which the Stage 1 diffusion mesh generator is replaced by a flow-matching mesh generator, while Stage 2 remains the same diffusion field predictor. Table 5 reports the resulting accuracy--runtime trade-off. Flow matching method reduces the total inference time from 315.1 ms to 169.5 ms, but also increases the prediction error, indicating that faster mesh generation is feasible but currently trades speed for accuracy. Reducing this trade-off remains an important direction for future work. More
advanced flow-matching or consistency-model parameterizations may further reduce sampling cost without sacrificing discretization
quality.

\begin{table}[htbp]
\centering\renewcommand{\arraystretch}{1.12}
\begin{tabular}{l c c c c}
\hline
Dataset & Mesh generator & Rel L2 & Spec L2 & Total time ms \\
\hline
Burgers & Diffusion & 0.0083 & 0.0072 & 315.1 \\
Burgers & Flow matching & 0.0180 & 0.0133 & 169.5 \\
\hline
\end{tabular}
\caption{Accuracy and runtime comparison between the default Stage 1 diffusion mesh generator and a flow-matching mesh generator on Burgers at $N=64$. Stage 2 is kept as the same diffusion field predictor.}
\label{tab:flow_matching_runtime}
\end{table}

\subsection{Mesh-Generation Pareto Analysis}
\begin{figure*}[htbp]
\centering
\includegraphics[width=0.98\textwidth]{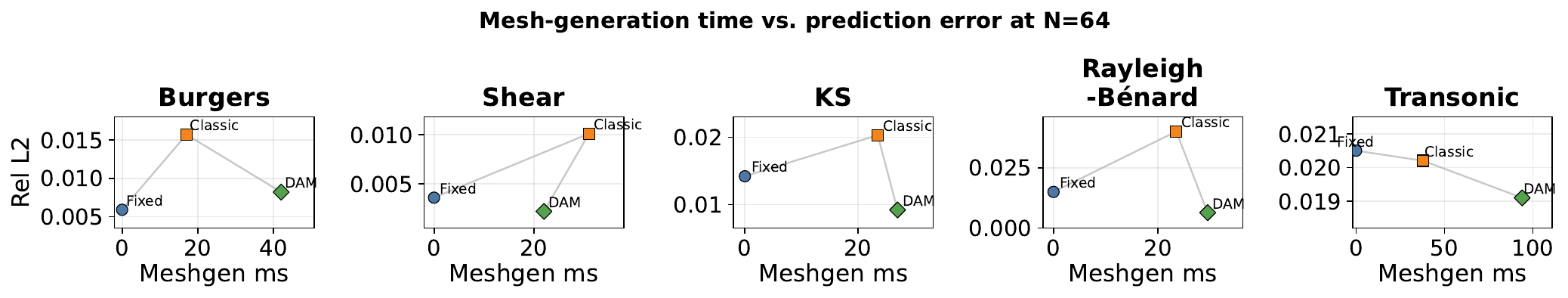}
\caption{Mesh-generation Pareto plot. Rel L2 is compared against mesh-generation time at resolution of $N=64$.}
\label{fig:meshgen_pareto}
\end{figure*}

Figure~\ref{fig:meshgen_pareto} compares fixed, classical adaptive, and DAM discretization policies using mesh-generation time as the horizontal axis and Rel L2 as the error axis. This isolates the cost of producing the discretization itself, while Table~\ref{tab:runtime_cost} reports DAM's total inference time.

The plot shows that the benefit of paying for learned mesh generation is regime-dependent. On KS and Periodic shear, DAM moves toward lower error without increasing mesh-generation cost relative to the classical policy. On Rayleigh-B\'{e}nard and Transonic, DAM obtains the best error but spends more mesh-generation time, so it should be read as an accuracy-capacity trade-off. Burgers is the boundary case: the fixed grid already aligns well with the 1D transport structure, so extra learned mesh generation does not dominate the fixed control. This supports the paper's narrower claim that discretization should be learned when the regime benefits from adaptive allocation, not that learned meshes are universally cheaper.

\subsection{Baseline Models}
All adaptive baselines are evaluated as discretization policies attached to the same FNO forecasting solver and the same train/validation/test splits. This is an adaptation of their mesh or sampling mechanisms rather than a claim that every original paper used an identical solver stack. We use this protocol deliberately: it isolates whether the adaptive allocation policy helps under a common forecasting backend, instead of confounding mesh adaptation with different PDE solvers, FEM implementations, or task-specific simulators.
\begin{itemize}
\item \textbf{Text2PDE}~\cite{zhou2024text2pde}: a latent diffusion framework built around mesh-aware autoencoding and conditional generation. We use it as the reduced-order generative baseline because it changes the state representation while keeping the online discretization policy separate from our learned mesh sampler.

\item \textbf{ATS}~\cite{fayyaz2022ats}: Adaptive Token Sampling was originally proposed for efficient vision transformers. We adapt its selection principle as a density-conditioned sampling baseline: informative locations receive higher sampling density before being passed to the same forecasting protocol. It is therefore a sampling-adaptation baseline rather than a classical mesh generator.

\item \textbf{AMR-MGNN}~\cite{perera2024multiscale}: a multiscale graph neural network coupled with adaptive mesh refinement for mesh-based simulations. We adapt its multiscale graph/message-passing idea into our forecasting setup as an AMR-style adaptive-discretization baseline.

\item \textbf{AMBER}~\cite{freymuth2025amber}: an iterative sizing-field method for adaptive mesh generation from expert demonstrations. In our comparison, AMBER supplies a learned sizing field that drives node-density allocation, followed by the shared prediction protocol.

\item \textbf{wLMR}~\cite{ozgenxian2020wavelet}: a wavelet-based local mesh-refinement method originally designed for rainfall-runoff simulations. We adapt its wavelet slope/curvature indicator as a lightweight local refinement baseline.
\end{itemize}

\subsection{Dataset Information}
\begin{itemize}
\item \textbf{Burgers}~\cite{kossaifi2025librarylearningneuraloperators}: a nonlinear advection-diffusion benchmark generated with the neural operator data utilities, used as the easiest low-dimensional transport regime.

\item \textbf{Periodic shear}~\cite{burns2020dedalus}: a periodic shear-flow benchmark generated with Dedalus, emphasizing smooth coherent shear structures and testing whether adaptive discretization avoids unnecessary distortion.

\item \textbf{KS}~\cite{shysheya2024conditional}: the chaotic Kuramoto-Sivashinsky benchmark, characterized by broadband oscillations and sensitivity to high-frequency structure.

\item \textbf{Rayleigh-B\'{e}nard}~\cite{burns2020dedalus}: a convection-style benchmark generated with Dedalus, used to test allocation around cellular transport patterns.

\item \textbf{Transonic}~\cite{kohl2026benchmarking}: a compressible-flow benchmark with shock-like discontinuities and sharp geometric transitions.
\end{itemize}

All datasets are converted to the same split-based HDF5 interface used by the implementation. The train split is used to fit the model and compute normalization statistics, the validation split is used for model selection in baseline training, and the test split is reserved for final Rel L2, RMSE, spectral error, mesh-quality, and cross-resolution metrics. In all experiments, the model receives a short history of consecutive fields and predicts the next field at the next time step. The benchmark suite is ordered roughly from easier to harder regimes in Table~\ref{tab:results}, so the comparison tests whether learned discretization remains useful beyond a single PDE family.

\subsection{Full Mesh-Quality Audit}
Prediction accuracy alone is insufficient for evaluating adaptive discretization: a low forecasting error is of limited value if the generated mesh exhibits element inversion, poor conditioning, or large reconstruction artifacts. Table~\ref{tab:mesh_validity_efficiency_full} complements the prediction results with a geometric audit. Across all datasets, DAM consistently maintains positive minimum Jacobians, minimum angles near $45^\circ$, and aspect ratios close to 2, indicating an orientation-preserving and well-conditioned mesh. In contrast, several adaptive baselines produce meshes with negative minimum Jacobians, extremely small minimum angles, or highly elongated elements, even when their reconstruction or prediction errors remain competitive. This gap should be interpreted in light of the evaluation protocol: these baselines originate as simulation/adaptation mechanisms and are converted here into moving triangular-mesh policies under a common neural forecasting interface. Poor element quality therefore mainly limits their engineering usability rather than automatically invalidating their field prediction numbers; a small Cycle Rel L2 can still preserve tensor reconstruction well enough for the downstream predictor. This is precisely why the geometric audit is useful: it separates predictive accuracy from mesh validity and shows that DAM obtains competitive prediction while retaining the validity constraints needed for a deployable adaptive discretization.

\begin{table*}[htbp]
\centering\renewcommand{\arraystretch}{1.10}
\begin{tabular}{l l c c c c}
\hline
Dataset & Method & Cycle Rel L2 $\downarrow$ & Min angle $\uparrow$ & Aspect $\downarrow$ & Min $J$ $\uparrow$ \\
\hline
Burgers
& ATS & $0.272\pm0.023$ & $0.46$ & $131.19$ & $-3.82{\times}10^{-1}$ \\
& AMR-MGNN & $0.303\pm0.031$ & $0.36$ & $177.84$ & $-4.18{\times}10^{-1}$ \\
& AMBER & $0.376\pm0.031$ & $1.65$ & $124.53$ & $-1.89$ \\
& wLMR & $0.538\pm0.061$ & $0.08$ & $779.32$ & $-5.07$ \\
& DAM (Ours) & $\mathbf{0.266\pm0.026}$ & $\mathbf{44.81}$ & $\mathbf{2.00}$ & $\mathbf{2.40{\times}10^{-3}}$ \\
\hline
Shear
& ATS & $0.129\pm0.004$ & $0.79$ & $72.24$ & $-3.47{\times}10^{-1}$ \\
& AMR-MGNN & $0.131\pm0.005$ & $0.79$ & $73.02$ & $-3.49{\times}10^{-1}$ \\
& AMBER & $0.135\pm0.005$ & $0.56$ & $103.56$ & $-5.66{\times}10^{-1}$ \\
& wLMR & $0.139\pm0.006$ & $0.46$ & $125.74$ & $-7.93{\times}10^{-1}$ \\
& DAM (Ours) & $\mathbf{0.128\pm0.005}$ & $\mathbf{44.82}$ & $\mathbf{2.00}$ & $\mathbf{2.40{\times}10^{-3}}$ \\
\hline
KS
& ATS & $0.146\pm0.012$ & $0.65$ & $89.77$ & $-3.61{\times}10^{-1}$ \\
& AMR-MGNN & $0.148\pm0.011$ & $0.61$ & $97.44$ & $-3.64{\times}10^{-1}$ \\
& AMBER & $0.219\pm0.016$ & $0.17$ & $336.98$ & $-2.51$ \\
& wLMR & $0.213\pm0.023$ & $0.12$ & $490.45$ & $-2.77$ \\
& DAM (Ours) & $\mathbf{0.144\pm0.009}$ & $\mathbf{44.78}$ & $\mathbf{2.01}$ & $\mathbf{2.40{\times}10^{-3}}$ \\
\hline
Rayleigh
& ATS & $0.260\pm0.021$ & $0.72$ & $79.86$ & $-4.05{\times}10^{-1}$ \\
-B\'{e}nard& AMR-MGNN & $0.261\pm0.018$ & $0.75$ & $76.90$ & $-4.00{\times}10^{-1}$ \\
& AMBER & $\mathbf{0.260\pm0.022}$ & $0.62$ & $108.09$ & $-8.43{\times}10^{-1}$ \\
& wLMR & $0.271\pm0.023$ & $0.57$ & $145.84$ & $-9.04{\times}10^{-1}$ \\
& DAM (Ours) & $0.261\pm0.024$ & $\mathbf{44.82}$ & $\mathbf{2.00}$ & $\mathbf{2.40{\times}10^{-3}}$ \\
\hline
Transonic
& ATS & $0.182\pm0.006$ & $0.71$ & $81.67$ & $-3.63{\times}10^{-1}$ \\
& AMR-MGNN & $0.187\pm0.006$ & $0.57$ & $100.12$ & $-4.15{\times}10^{-1}$ \\
& AMBER & $0.441\pm0.022$ & $1.80$ & $33.79$ & $2.46{\times}10^{-4}$ \\
& wLMR & $0.287\pm0.013$ & $0.62$ & $94.39$ & $6.44{\times}10^{-5}$ \\
& DAM (Ours) & $\mathbf{0.179\pm0.005}$ & $\mathbf{44.65}$ & $\mathbf{2.00}$ & $\mathbf{2.39{\times}10^{-3}}$ \\
\hline
\end{tabular}
\caption{Mesh-quality comparison at resolution of $N=64$ across adaptive-discretization methods, recomputed using the same held-out sampling protocol. Cycle Rel L2 measures interpolation/reconstruction error when moving between adaptive and regular representations, Min angle and aspect ratio measure element conditioning, and Min $J$ is the minimum signed element area/Jacobian proxy. Positive Min $J$ together with large minimum angles and aspect ratios near two indicates an orientation-preserving, well-conditioned moving mesh; small angles and large aspect ratios indicate sliver elements even when the signed area is nonnegative.}
\label{tab:mesh_validity_efficiency_full}
\end{table*}

\subsection{Future Work}
Several limitations of the current framework point to concrete extensions. The iterative denoiser used by the mesh branch is a direct source of additional inference cost. Since DAM currently generates mesh displacements through diffusion denoising, future work may replace this mesh generator with faster continuous generative parameterizations, such as flow matching, rectified flow, consistency models, or other low-step transport-based generators. The experiment in Table~\ref{tab:flow_matching_runtime} suggests that such replacements can reduce runtime, but preserving the accuracy and mesh quality of the diffusion-based discretization remains an open challenge.

Another direction is uncertainty analysis for generative discretization. Although Gaussian noise is injected during the diffusion process, the supervised forecasting objective and geometric regularization drive the mesh generator toward a stable task-specific mesh, so the observed stochastic variation in generated meshes is often small. This behavior means that mesh samples from the current model should not be interpreted as calibrated posterior uncertainty. The current paper therefore does not address uncertainty quantification. Future work could study whether explicitly uncertainty-aware objectives, ensembles, or noise-conditioned evaluation protocols can make stochastic mesh variation more informative under distribution shift, noisy observations, or multi-modal solution regimes.

\begin{table*}[htbp]
\centering\renewcommand{\arraystretch}{1.12}
\begin{tabular}{l c c c c}
\hline
Dataset & Error hotspot & $|\nabla u|$ & Vorticity & Cond. proxy \\
\hline
Burgers & $-0.017\pm0.171$ & $0.114\pm0.130$ & $0.155\pm0.164$ & $0.130\pm0.150$ \\
Shear & $-0.079\pm0.061$ & $0.020\pm0.099$ & $-0.002\pm0.083$ & $0.013\pm0.092$ \\
KS & $0.073\pm0.103$ & $0.085\pm0.135$ & $-0.104\pm0.103$ & $-0.036\pm0.124$ \\
Rayleigh-B\'{e}nard & $0.138\pm0.067$ & $0.249\pm0.068$ & $-0.008\pm0.077$ & $0.097\pm0.065$ \\
Transonic & $0.179\pm0.044$ & $0.178\pm0.059$ & $0.071\pm0.051$ & $0.159\pm0.053$ \\
\hline
\end{tabular}
\caption{Spearman correlation between learned node density and interpretable structure maps across 24 held-out test draws. The strongest positive alignment appears in Transonic and Rayleigh-B\'{e}nard regimes, while Periodic shear remains weakly aligned because the field is already well represented by simple structure.}
\label{tab:mesh_semantics}
\end{table*}

\begin{figure}[htbp]
\centering
\includegraphics[width=0.94\columnwidth]{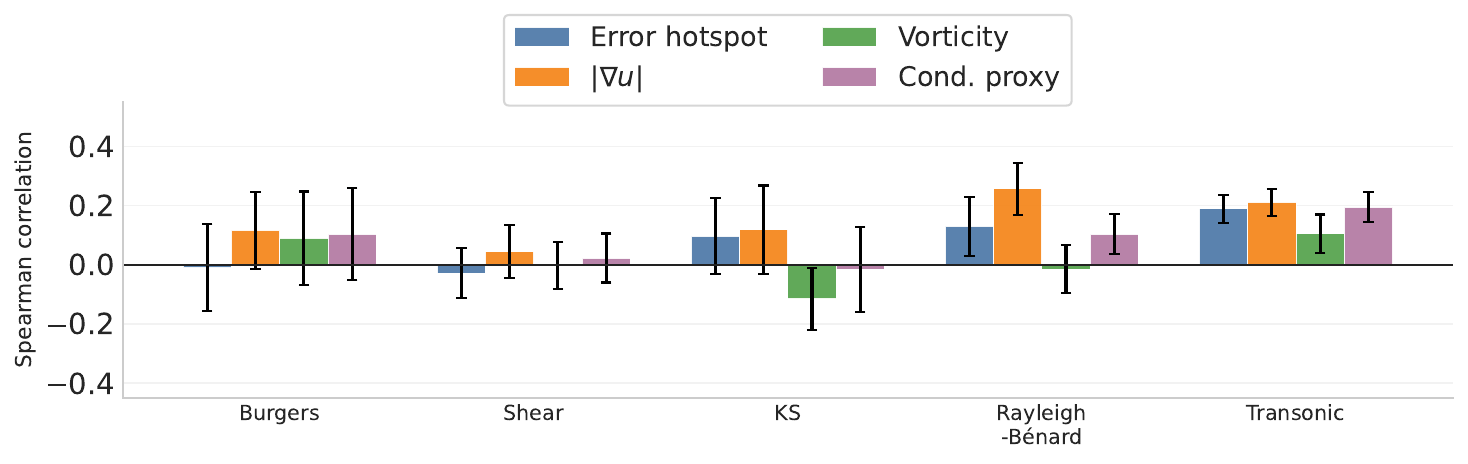}
\caption{Semantic-correlation audit. Bars report mean Spearman correlation with one-standard-deviation error bars.}
\label{fig:mesh_semantics}
\end{figure}

\subsection{What the Mesh Learns}
Beyond prediction error, we ask whether the learned mesh responds to an interpretable field structure. Table~\ref{tab:mesh_semantics} and Figure~\ref{fig:mesh_semantics} report a semantic-correlation audit between learned node density and error hotspots, gradient magnitude, vorticity, and the conditioning proxy. This matters because an adaptive mesh can move nodes without placing them near structures that govern forecasting error. We therefore treat node density as an observable consequence of the learned discretization policy and test whether it is rank-correlated with interpretable maps computed from held-out trajectories.

The result is useful precisely because it is not uniform. Transonic and Rayleigh-B\'{e}nard show the clearest positive alignment with localized spatial structure, where shocks, boundary-layer-like features, or coherent convection patterns make extra resolution meaningful. Periodic shear remains close to zero because its dominant pattern is simple and globally organized; in that case, strong local density correlation is neither expected nor necessary. Burgers shows moderate alignment with gradient and vortical proxies, and KS remains mixed, consistent with broadband dynamics where local monotone cues are weaker. The point is not that the same proxy explains every dataset, but that the learned mesh reveals different allocation behavior across regimes.

This complements the error tables: Table~\ref{tab:results} establishes forecasting competitiveness, Figures~\ref{fig:spectral_concentration}-\ref{fig:spectral_ablation} show spectral consequences, and the semantic audit asks whether allocation can be read in physical space. Together, these results support the narrower claim that diffusion-based discretization is a regime-dependent representation, not a fixed refinement template. They also clarify why diffusion is a natural parameterization: the mesh is generated as a structured object under validity and spectral constraints, rather than selected from a single handcrafted rule.

\subsection{Parameter Generalization Audit}
To test whether the learned discretization remains usable under parameter shifts, we evaluate the trained DAM checkpoints without retraining on harder Periodic-shear, Rayleigh-B\'{e}nard, and Transonic test cases. The in-distribution Periodic-shear training data use the default $Re\approx500$ setting, and the audit raises it to $Re\in\{2000,5000,10000\}$. The Rayleigh-B\'{e}nard training data use the default $Ra\approx1.6\times10^5$ setting, and the audit evaluates stronger convection at $Ra\in\{1.6\times10^6,4\times10^6,8\times10^6\}$. The Transonic training set covers Mach numbers between $0.7$ and $0.9$, and the audit evaluates higher-Mach test cases $M\in\{1.0,1.1,1.2\}$. Even under these shifts, Table~\ref{tab:param_generalization} shows smooth degradation rather than catastrophic failure; the largest sensitivity appears in the Transonic tests, where both field and spectral errors increase monotonically with Mach number.

\begin{table}[htbp]
\centering\renewcommand{\arraystretch}{1.28}
\begin{tabular}{l c c c c}
\hline
Regime & Parameter & Rel L2 $\downarrow$ & RMSE $\downarrow$ & Spec L2 $\downarrow$ \\
\hline
Shear & Re=2000 & 0.0031 & 0.0015 & 0.0012 \\
 & Re=5000 & 0.0031 & 0.0015 & 0.0013 \\
 & Re=10000 & 0.0031 & 0.0016 & 0.0013 \\
\hline
Rayleigh-B\'{e}nard & Ra=$1.6\times10^6$ & 0.0164 & 0.0022 & 0.0151 \\
 & Ra=$4\times10^6$ & 0.0094 & 0.0014 & 0.0051 \\
 & Ra=$8\times10^6$ & 0.0160 & 0.0025 & 0.0222 \\
\hline
Transonic & M=1.0 & 0.0242 & 0.0147 & 0.0043 \\
 & M=1.1 & 0.0300 & 0.0201 & 0.0105 \\
 & M=1.2 & 0.0373 & 0.0273 & 0.0206 \\
\hline
\end{tabular}
\caption{Parameter generalization audit for DAM (Ours) at the resolution of $N=64$ without retraining. Periodic-shear and Rayleigh-B\'{e}nard rows are generated by harder physical parameters. Transonic rows report cross-Mach-number evaluation beyond the estimated training range.}
\label{tab:param_generalization}
\end{table}

The parameter-shift audit separates three behaviors. Periodic shear remains almost unchanged even up to $Re=10000$, consistent with its globally organized shear structure. Rayleigh-B\'{e}nard is less monotone: the intermediate convection setting is easiest, while both weaker and stronger convection increase spectral error. Transonic shows the clearest extrapolation trend: increasing Mach number steadily raises Rel L2, RMSE, and spectral error, indicating that the learned discretization remains usable while the solution branch becomes more sensitive to compressive structures beyond the training range.

\subsection{Cross-Resolution Quantitative Audit}
The learned mesh policy is trained at $N=64$ and evaluated without retraining at $N\in\{16,32,48,64,96,128,256\}$. Table~\ref{tab:cross_mesh_quant} complements Figure~\ref{fig:cross_resolution_main} with Rel L2 metrics across resolutions.
\begin{table*}[htbp]
\centering\renewcommand{\arraystretch}{1.12}
\begin{tabular}{l c c c c c c c}
\hline
Dataset & $N=16$ & $N=32$ & $N=48$ & $N=64$ & $N=96$ & $N=128$ & $N=256$ \\
\hline
Burgers & 0.0554 & 0.0081 & 0.0092 & 0.0082 & 0.0084 & 0.0089 & 0.0087 \\
Shear & 0.0045 & 0.0030 & 0.0030 & 0.0030 & 0.0031 & 0.0032 & 0.0032 \\
KS & 0.0363 & 0.0196 & 0.0103 & 0.0102 & 0.0111 & 0.0113 & 0.0109 \\
Rayleigh-Benard & 0.0846 & 0.0070 & 0.0083 & 0.0070 & 0.0079 & 0.0100 & 0.0086 \\
Transonic & 0.1035 & 0.0256 & 0.0198 & 0.0188 & 0.0191 & 0.0190 & 0.0192 \\
\hline
\end{tabular}
\caption{Cross-resolution Rel L2 of DAM (Ours) when evaluated at different mesh sizes without retraining. Each entry reports the mean Rel L2 over held-out evaluation batches.}
\label{tab:cross_mesh_quant}
\end{table*}

The quantitative trend is consistent with the visual audit. The model is trained at $N=64$, yet Burgers, Periodic shear, KS, Rayleigh-B\'{e}nard, and Transonic all remain close to their $N=64$ error once the evaluation mesh reaches moderate resolution. The main degradation appears at $N=16$, especially for Rayleigh-B\'{e}nard and Transonic, where localized convection or shock-sensitive structure requires enough nodes before the learned allocation can be useful. This supports the interpretation that DAM learns a resolution-aware discretization policy rather than memorizing a single grid size; the policy is not resolution-free, but it remains stable over a broad range around the training mesh.

\subsection{Supplementary Mesh-Size Visualization}
This final appendix section collects the field-level qualitative visualizations and cross-mesh-size visualizations used to audit prediction structure and resolution robustness.

\begin{figure}[htbp]
\centering
\includegraphics[width=0.78\textwidth]{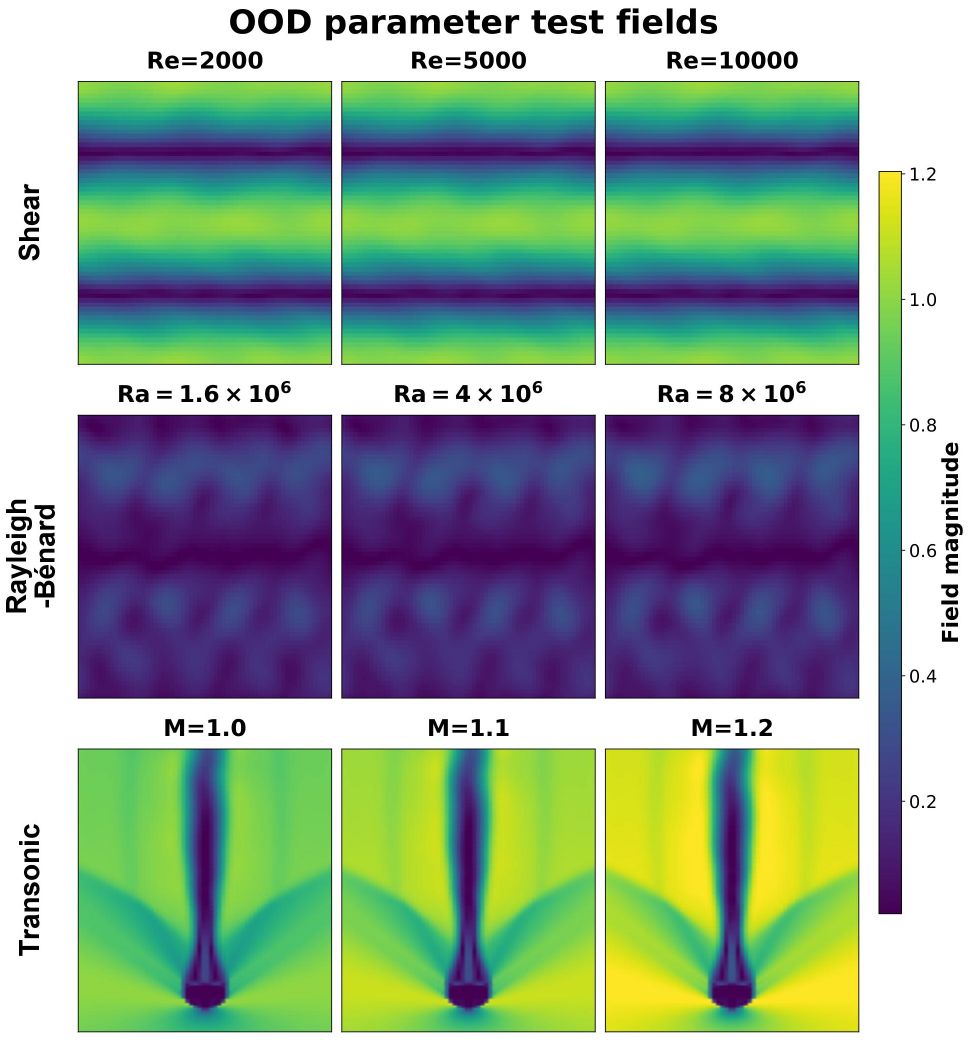}
\caption{Parameter-shift fields. Columns increase the physical parameter within each regime.}
\label{fig:param_generalization_fields}
\end{figure}

\begin{figure}[htbp]
\centering
\includegraphics[width=0.98\textwidth]{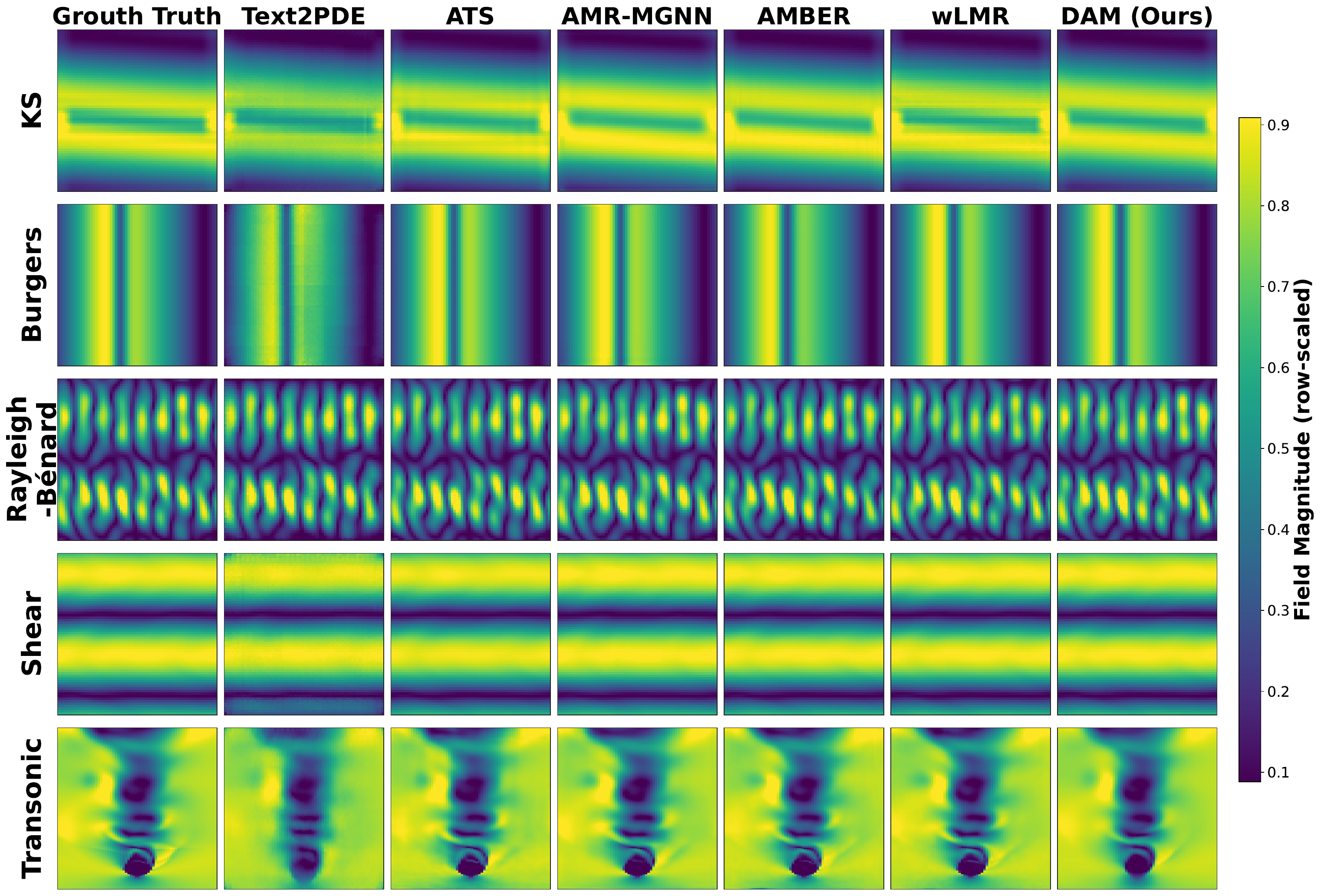}
\caption{Prediction heatmaps. Field-level predictions are compared across five datasets and methods.}
\label{fig:prediction_heatmap}
\end{figure}

\begin{figure}[htbp]
\centering
\includegraphics[width=0.98\textwidth]{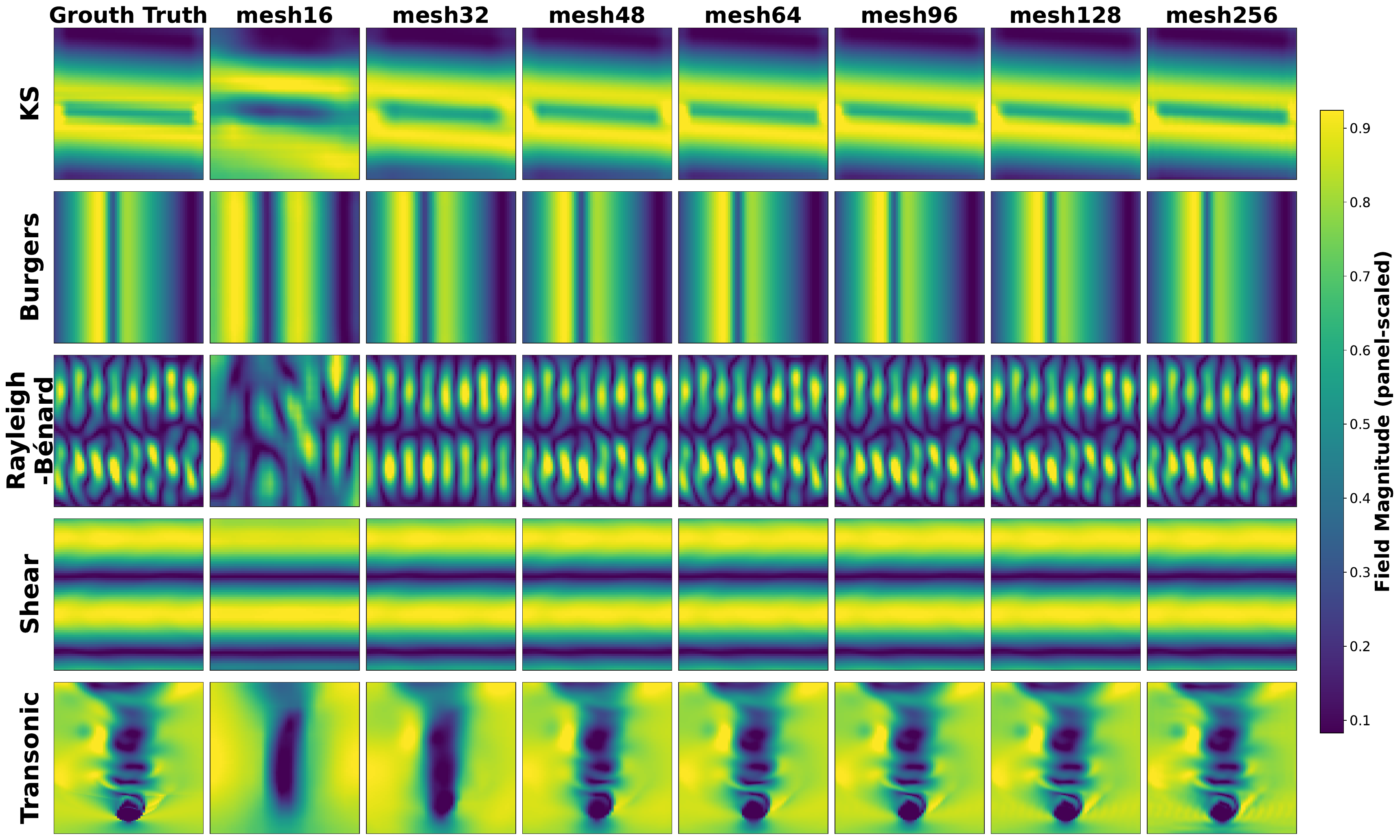}
\caption{Cross-resolution heatmaps. DAM is trained at $N=64$ and evaluated without retraining across multiple mesh sizes.}
\label{fig:cross_resolution_main}
\end{figure}

\end{document}